\documentclass[10pt,twocolumn,letterpaper]{article}

\usepackage{cvpr}              
\usepackage{indentfirst}
\usepackage[accsupp]{axessibility}
\usepackage{makecell}
\usepackage{booktabs}
\usepackage{multirow}
\usepackage{colortbl}
\usepackage{xcolor}
\usepackage{array}
\usepackage{float}
\usepackage[ruled,vlined]{algorithm2e}
\usepackage[dvipsnames]{xcolor}
\usepackage{times}  
\usepackage[T1]{fontenc}  

\definecolor{commentcolor}{HTML}{d62c2c} 
\definecolor{forestgreen}{HTML}{009B55} 
\definecolor{myred}{rgb}{ .753,  0,  0}
\definecolor{shallowgray}{HTML}{738385}

\definecolor{commentcolor_py}{RGB}{110,154,155}   
\definecolor{keycolor}{rgb}{0.858, 0.188, 0.478} 

\RequirePackage[breakable,skins]{tcolorbox}

\definecolor{forestgreen}{HTML}{009B55} 
\definecolor{myred}{rgb}{.753,  0,  0}
\definecolor{shallowgray}{HTML}{738385}
\definecolor{colorcommentfg}{RGB}{0,63,87}
\definecolor{colorcommentbg}{HTML}{e0f0f6}
\definecolor{colorcommentframe}{RGB}{0,112,155}

\colorlet{colorchangebg}{black!2}
\colorlet{colorchangeframe}{black!20}

\definecolor{cvprblue}{rgb}{0.21,0.49,0.74}
\definecolor{myblue}{RGB}{210, 234, 255}
\definecolor{myred}{RGB}{255, 210, 210}

\usepackage[pagebackref,breaklinks,colorlinks,allcolors=cvprblue]{hyperref}

\def\paperID{4149} 
\def\confName{CVPR}
\def\confYear{2025}

\title{Knowledge Bridger: Towards Training-Free Missing Modality Completion}

\author{
Guanzhou Ke$^{1,2}$, Shengfeng He$^{2,}$\thanks{Corresponding authors (Email: shengfenghe@smu.edu.sg, wangbo@ia.ac.cn).}~, Xiaoli Wang$^6$, Bo Wang$^{4,\textasteriskcentered{}}$, \\Guoqing Chao$^5$, Yuanyang Zhang$^3$, Yi Xie$^7$, Hexing Su$^8$\\
{\hspace{-0.4cm}\normalsize $^1$Beijing Jiaotong University, $^2$Singapore Management University, $^3$Southeast University,}  \\
{\hspace{-0.4cm}\normalsize $^4$Institute of Automation, Chinese Academy of Sciences, $^5$Harbin Institute of Technology,} \\
{\hspace{-0.4cm}\normalsize $^6$Nanjing University of Science and Technology, $^7$South China University of Technology, $^8$Xiamen Institute of Technology} \\
}

\begin{document}
\maketitle


\begin{abstract}
Previous successful approaches to missing modality completion rely on carefully designed fusion techniques and extensive pre-training on complete data, which can limit their generalizability in out-of-domain (OOD) scenarios. In this study, we pose a new challenge: \textbf{can we develop a missing modality completion model that is both resource-efficient and robust to OOD generalization?} To address this, we present a training-free framework for missing modality completion that leverages large multimodal model (LMM). Our approach, termed the ``Knowledge Bridger'', is modality-agnostic and integrates generation and ranking of missing modalities. By defining domain-specific priors, our method automatically extracts structured information from available modalities to construct knowledge graphs. These extracted graphs connect the missing modality generation and ranking modules through the LMM, resulting in high-quality imputations of missing modalities. Experimental results across both general and medical domains show that our approach consistently outperforms competing methods, including in OOD generalization. Additionally, our knowledge-driven generation and ranking techniques demonstrate superiority over variants that directly employ LMMs for generation and ranking, offering insights that may be valuable for applications in other domains. Our code is accessible at: \url{https://github.com/Guanzhou-Ke/Knowledge-Bridger}.
\end{abstract}    


\section{Introduction}
\label{sec:intro}

Multimodal learning \cite{ngiam2011multimodal, jiang2024delving} is a foundational task in artificial intelligence that enables models to integrate diverse data types—such as text, images, and audio—to enhance their representational and reasoning capabilities \cite{ren2021learning, lv2021differentiated, liu2025unsupervised}. However, real-world applications often encounter incomplete or missing modalities due to factors like noise, sensor failures, or privacy constraints, which can compromise a model’s robustness and generalizability. Missing modality completion (MMC) addresses this issue by imputing or reconstructing absent modalities based on available data, allowing models to fully leverage multimodal information. Recent MMC methods \cite{guo-etal-2024-multimodal, wang2023distribution, lee2023multimodal, cai2018deep} have demonstrated promising results across various applications. For instance, in medical diagnostics \cite{zhang2022m3care, cohen2023joint}, MMC models can reconstruct absent diagnostic data (e.g., MRI, CT, and X-rays) by leveraging existing medical reports or related images.

Despite these advancements, MMC methods often have limited out-of-domain (OOD) transferability, typically requiring retraining on new domain data to maintain performance. For example, some MMC models \cite{zhao2021missing, ma2021smil} developed for sentiment analysis need substantial adaptations to be effective in applications like medical imaging or autonomous driving, leading to significant human and computational costs. To address these challenges, recent works have proposed domain-agnostic solutions, such as missing modality tags \cite{zeng2022tag} to aid in predicting absent modalities or prompt-learning techniques \cite{guo-etal-2024-multimodal, lee2023multimodal} that dynamically adjust fusion strategies. While these methods lower domain adaptation costs, they still require extensive training data, limiting their effectiveness in data-scarce domains, such as rare disease analysis. This raises a crucial question: \textit{Can we develop an MMC model that achieves both low-resource dependency (e.g., minimal computational and domain-specific data requirements) and strong OOD capability?}

Recent advancements \cite{wang2024qwen2, liu2024visual, team2024chameleon, hurst2024gpt} in large multimodal models have highlighted their strong OOD capabilities and adaptability to new tasks with minimal resources \cite{kojima2022large}. In this work, we explore the potential of leveraging LMM to effectively address the MMC challenge. Before progressing, it is essential to distinguish between modality generation and missing modality completion. Modality generation typically focuses on creating new modalities from available information, prioritizing \textbf{diversity and creativity} in the generated content. In contrast, missing modality completion emphasizes \textbf{accuracy} over diversity, reconstructing absent modalities to maintain semantic coherence and improve task performance. 

Addressing MMC with LMM presents two primary challenges: \textit{1) \textbf{generation}: applying constraints on modality generation to ensure fidelity to the missing content}, and \textit{2) \textbf{ranking}: selecting the most appropriate completion from generated candidates}. Effective use of LMM for these tasks requires embedding sufficient domain knowledge to guide accurate interpretation and reconstruction of missing content. For example, generating an accurate image in a vision-language task can be challenging with only brief textual descriptions; however, incorporating detailed descriptions of entities and their interactions allows LMM to more accurately reconstruct missing data. Additionally, such prior knowledge aids LMM in ranking candidate completions by identifying the most semantically plausible options. Leveraging LMM’s in-context learning capability \cite{wei2022emergent}, we can non-intrusively embed prior knowledge, such as using the Chain-of-Thought (CoT) approach \cite{wei2022chain}, to enhance the model's understanding of missing modalities.

Based on these insights, we propose a novel, training-free MMC approach, termed ``Knowledge Bridger'', which autonomously mines multimodal knowledge, generates missing modalities, and ranks the best completions. This method comprises three main modules: a knowledge modeling module, a knowledge-driven modality generation module, and a ranking module. In the knowledge modeling stage, we employ LMM to analyze available modalities and extract key elements such as objects, interactions, and attributes using the CoT approach. For specialized fields like medical imaging, we use in-context learning to integrate domain-specific knowledge, reducing reliance on extensive target domain data. Following knowledge extraction, we construct a knowledge graph to represent the relationships and attributes of available and missing modalities, providing LMM with a structured reference for generating missing content. To ensure high-quality completion, the knowledge-driven generation module utilizes this graph to guide LMM in generating specific content with precise details, such as object locations and observable attributes. Finally, the ranking module, informed by expert knowledge, assesses each generated completion by computing graph and representational similarity scores. Specifically, a graph similarity score is derived from comparing the knowledge graphs of the available and generated modalities, while a representational similarity score is calculated using models like BLIP \cite{li2023blip} and CLIP \cite{radford2021learning}. A weighted average of these scores offers the final assessment. 

Our extensive experiments indicate that our method markedly improves MMC performance in both general and OOD scenarios. We also find that our approach scales effectively with LMM, with larger models yielding higher-quality completions. For instance, using OpenAI's GPT-4o \cite{hurst2024gpt} results in a significant performance boost across metrics compared to models with 72B or 7B parameters. The visualization results show that our method outperforms the conditional generation variant in generating missing modalities. Our contributions are summarized as follows:

\begin{itemize}
\item We introduce a training-free pipeline to address missing modality completion, leveraging LMM to automatically extract multimodal knowledge, generate missing modalities, and rank completions. To our knowledge, this is the first work to apply LMM to MMC tasks.
\item We delve into a modality-agnostic unified strategy for both missing modality completion and ranking. This approach allows us to focus on defining domain-specific knowledge without the necessity of intricate fusion methods or implementing a specialized training pipeline. 
\item We present extensive experimental evidence demonstrating that our method facilitates domain transfer, outperforming other MMC methods in both general and OOD scenarios. Additionally, our generated completion data improves the performance of other MMC models.
\end{itemize}

\section{Method}
\label{sec:method}

\subsection{Overview}

\begin{figure*}[h]
  \centering
   \includegraphics[width=1\linewidth]{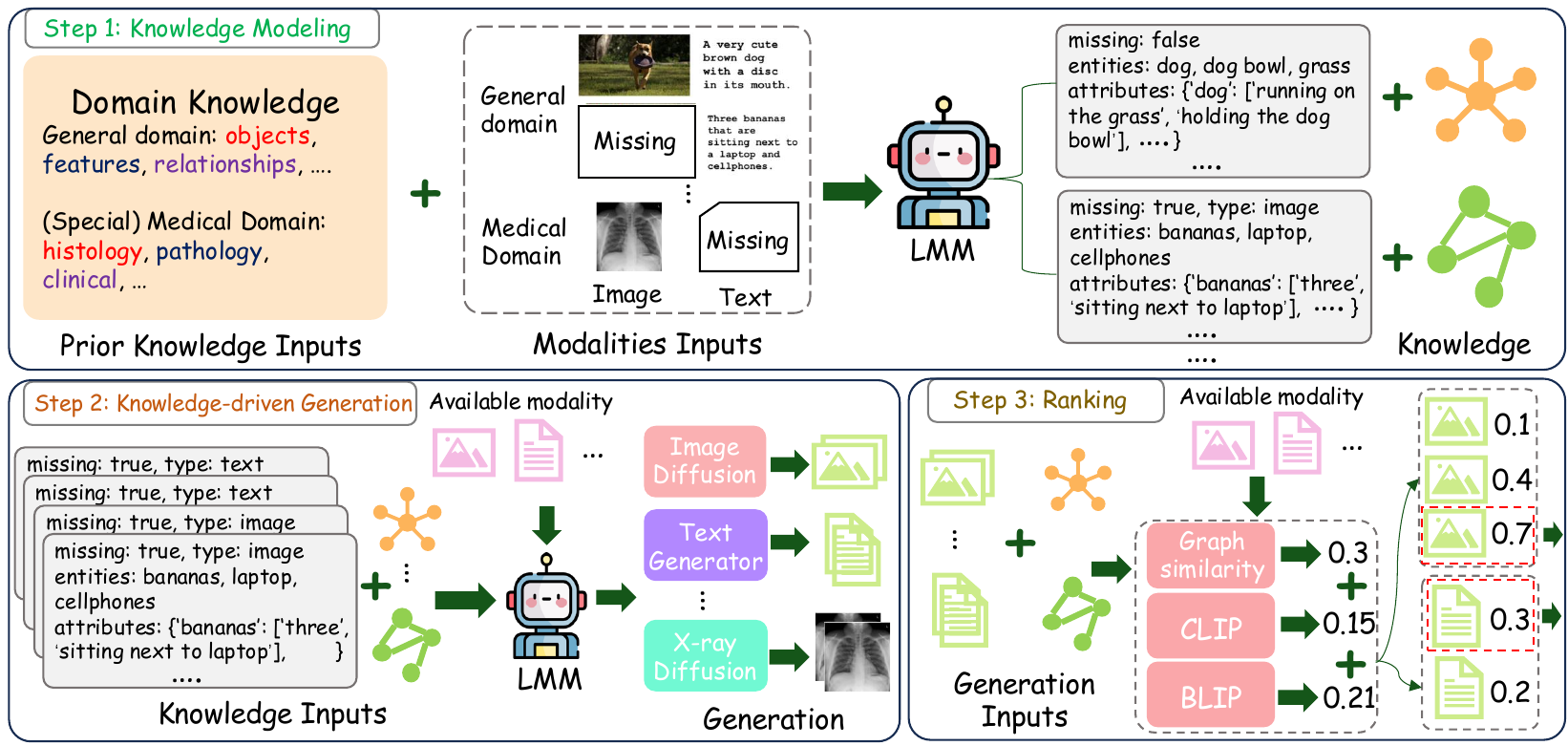}
   \caption{\textbf{Overview of `Knowledge Bridger' Pipeline.} The pipeline consists of three steps: 1) construction of a knowledge graph from available modalities, 2) knowledge-driven generation of missing modalities, and 3) knowledge-based ranking. The input of this pipeline is the available multimodal data and known domain knowledge. The output is the required missing modality.}
   \label{fig:framework}
\vspace{-10pt}
\end{figure*}

Our approach aims to construct a training-free MMC pipeline by leveraging a pre-trained LMM. This pipeline extracts and models knowledge from available modalities, and subsequently uses this knowledge to generate missing modality data and select the most appropriate generated candidate. The pipeline is depicted in Fig. \ref{fig:framework} and consists of three steps:

\begin{itemize}
    \item \textbf{Step 1: construction of a knowledge graph from available modalities} (in section \ref{sec:kg-modeling}). The goal of this step is to utilize the general knowledge a pre-trained LMM to understand the content of each modality and their interrelationships.
    \item \textbf{Step 2: knowledge-driven generation} (in section \ref{sec:kg-generation}). In this step, the LMM employs the knowledge graph to extract specific details about the missing modality, including the number and attributes of objects. A corresponding modality generator then uses this information to generate the required missing modality. 
    \item \textbf{Step 3: knowledge-based ranking} (in section \ref{sec:kg-ranking}). This step aims to compute a quality score for the generated missing data by assessing the graph similarity and semantic similarity between the missing and available modalities.
\end{itemize}

\subsection{Knowledge Graph Modeling}
\label{sec:kg-modeling}

Our objective is to automatically extract knowledge from available modalities to support the generation and ranking of any missing modalities. \textbf{In our context, `knowledge' refers to information encapsulating the characteristics of existing modalities, enabling the generation and ranking modules to create semantically consistent missing data.} Extracting relevant knowledge from unknown domains, however, presents significant challenges. From a knowledge graph perspective, constructing a meaningful, modality-specific graph requires predefined nodes and relationships. In a training-free context, predefining these elements becomes particularly challenging. 

To overcome this, we develop an automatic entity and relationship mining strategy using an LMM. This strategy leverages the extensive prior knowledge and OOD capabilities of LMM to identify entities and relationships across various modalities, even without predefined elements. Recent study \cite{kojima2022large} highlight LMM’s potential for zero-shot learning and reasoning.

Building on this analysis, LMM can be utilized to extract elements from available modalities using prompts. To enhance scalability, we propose the following extraction rule: \textit{\{Entity: Reasoning Prompts\}}. For instance, to identify potential objects, we could use: \textit{\{`Objects': `Identify the major objects in the [modality-type].'\}}. This rule allows us to incorporate object relationships and interaction data\footnote{The complete rules and prompts are provided in the appendix.}. To improve adaptability across domains, domain-specific prior knowledge—such as histological and clinical diagnostic information for medical image analysis—can be included. This approach offers two main advantages: 1) reducing misconceptions when LMM operates in a new domain, and 2) enhancing its reasoning capabilities for novel entities.

LMM can further consolidate the extracted information into a modality-specific knowledge graph. A straightforward approach is to guide LMM to extract potential entity-relation pairs from the collected data. However, this method is constrained by the context window length, and an excessive number of extraction rules may lead to overlooked entity-relation pairs. To mitigate these limitations, we implement the Chain-of-Thought (CoT) method. Specifically, LMM is first directed to produce concise responses for each rule, followed by extracting unique entity-relation pairs from these responses. This stepwise decomposition improves both the accuracy of responses for each rule and the synthesis of information. Importantly, we extract and retain data solely from current modalities to avoid interference from unrelated information, thereby enhancing LMM’s reasoning efficiency.


\subsection{Knowledge-driven Generation}
\label{sec:kg-generation}

The goal of missing multimodal generation is to understand the content in available modalities and generate missing ones that align semantically. Two critical factors impact the quality of the generated missing modalities: understanding multimodal content and maintaining consistency. We previously discussed using LMM for content comprehension and knowledge graph extraction. Here, we explore using LMM for ensuring consistency and guiding generation. For convenience, we use image-text pairs as our study objects. For example, when an image is available, we aim to generate text that closely matches real data. A basic approach is using LMMs to describe the image directly. However, this method involves significant randomness. First, the form of the missing text is unknown—it could be a caption, summary, or description. Second, we cannot precisely specify the subject of the missing text.

To solve this, we propose a knowledge-driven entity alternation strategy. Using domain knowledge and extracted knowledge graphs, we select related entities.For instance, if missing data focuses on an entity such as an `\textit{objection},' we traverse elements within the knowledge graph related to `\textit{objection}.' We then adopt a multi-view generation manner, allowing the LMM to generate missing information with each element as the subject, while encompassing all nodes and attributes within the knowledge graph. These outputs are stored as standardized text descriptions, reducing randomness, enhancing result retrieval, and offering better control and interpretability. With these descriptions, modality generators can create missing data. For missing images, entity-based descriptors can guide conditional diffusion methods. For missing text, LMM process these descriptions to generate outputs. This approach works across various fields, using mature generation models with domain knowledge to create needed data without extra training. However, relying only on this may not guarantee full accuracy, which we discuss in the next section.

\subsection{Knowledge-based Ranking}
\label{sec:kg-ranking}

To achieve automatic ranking of generated missing data based on the provided knowledge, we introduce graph similarity and representation similarity. Graph similarity is computed as the average cosine similarity score of the adjacency matrices of two graphs, as presented in the following equation:

\begin{equation}
\label{eq:graph-cos}
    \cos_{graph}(A_{i}, B_{i}) = \frac{1}{n} \sum_{i=1}^{n} \frac{A_{i} \cdot B_{i}}{\|A_{i}\| \|B_{i}\|},
\end{equation}
where $A_i$ and $B_i$ represent the \( i \)-th row vectors of adjacency matrices $A$ and $B$, respectively. $\|A_i\|$ refers to the Euclidean norm of the \( i \)-th row: $\|A_i\| = \sqrt{\sum_{j=1}^{m} A_{ij}^2}$. $n$ and $m$ represent the rows and columns of the adjacency matrix, respectively. This metric reflects the degree of similarity between the two graphs, and its value is normalized between 0 and 100. On the other hand, we compute the representation similarity between the generated and available modalities to reflect semantic consistency. For consistency, we similarly employ cosine similarity to compute the similarity between two representations, expressed as $\cos(\mathbf{a}, \mathbf{b}) = \frac{\mathbf{a} \cdot \mathbf{b}}{\|\mathbf{a}\| \|\mathbf{b}\|}$, where $\mathbf{a}$ and $\mathbf{b}$ are the vectors of two modalities. Inspired by \cite{bianco2023improving}, we directly utilize CLIP \cite{radford2021learning} and BLIP \cite{li2023blip} to obtain semantic embeddings for each modality. Finally, we derive the following equation to compute the generation quality score between any pair of available and missing modalities:
\begin{equation}
\begin{aligned}
    \label{eq:qs-score}
     QS&(x_a, x_m)  = \cos_{graph}(f_a(x_a), f_a(x_m))  \\
    & + [\cos(f_c(x_a), f_c(x_m)) + \cos(f_b(x_a), f_b(x_m))],
\end{aligned}
\end{equation}
where $x_a$ and $x_m$ represent available and missing modalities, respectively. The functions $f_a(\cdot)$, $f_c(\cdot)$, and $f_b(\cdot)$ are used to obtain the adjacency matrix, CLIP's embedding, and BLIP's embedding of the given modality, respectively. We argue that $QS(\cdot, \cdot)$ can comprehensively assess two critical factors: knowledge structure similarity and semantic consistency. A higher value indicates a higher quality of the generated missing modality. Our method ultimately outputs the generated missing modality with the highest score.
\section{Experiments}

\subsection{Setup}
\noindent \textbf{Baselines.} To validate the effectiveness of our proposed method, we selected several baseline methods, categorized into two types: imputation-based and non-imputation-based. Imputation-based methods aim to restore missing modalities by learning relationships among the representations within each modality; for this category, we choose MMIN \cite{zhao2021missing} and DiCMoR \cite{wang2023distribution} as baselines. In contrast, non-imputation-based methods bypass the need to restore missing modalities or predict their representations, allowing downstream tasks to be completed without these steps. We choose MPLMM \cite{guo-etal-2024-multimodal} and MPMM \cite{lee2023multimodal} as baselines for this category. Additionally, we introduce a simple baseline that completely removing missing data. This baseline utilizes the pre-trained CLIP model as multimodal backbones. Subsequently, the modality features are concatenated and fed into a classification head (a single-layer MLP) to ultimately obtain the probability for each category.

\noindent \textbf{Dataset.} To evaluate the domain transferability of our method, we employ two general domain datasets and one OOD dataset. The general domain datasets include COCO-2014 \cite{lin2014microsoft} and MM-IMDb \cite{arevalo2017gated}. COCO-2014 \cite{lin2014microsoft} is a large-scale vision-language dataset containing approximately 81$K$ training samples and 41$K$ validation and test samples, with objects classified into 80 categories. In our setup, we treat it as a multimodal multi-label classification dataset, using the validation set as the test set. MM-IMDb \cite{arevalo2017gated} is a large-scale movie genre classification dataset with around 25,000 movies from IMDb, each represented by its movie poster and plot summary, and annotated with 27 multi-label genre tags. In our setup, we divide the dataset into 18,160 samples for the training set and 7,799 samples for the test set. On the other hand, we employ the IU X-ray \cite{demner2016preparing} dataset as our OOD dataset. This dataset includes 6,674 training and 756 testing samples. We further expanded the original 14 observations into 105 finer-grained categories based on location and severity, creating a long-tail dataset.

\noindent \textbf{Evaluation Metrics.} For multimodal multi-label classification on three datasets, we report the performance of comparison methods using the macro F1-score and Average Precision (AP). We report the average result of five different random seeds. Additionally, we introduce the mean similarity score (denoted \textbf{SS}) to evaluate the generation quality of our method and other imputation-based methods. Specifically, we utilize the vision and text backbones of the pre-trained CLIP \cite{radford2021learning} to compute embeddings for all ground-truth and generated missing modalities. The cosine similarity between each pair of embeddings is then computed and averaged, with scores ranging from 0 to 100. A higher score indicates better generation quality.

\noindent \textbf{Implementation Details.} We employ Qwen2-VL-7B \cite{wang2024qwen2} as our default large multimodal model\footnote{More implementation details are presented in the appendix.}. It is specifically designed to handle both visual and textual inputs, enabling it to process and interpret complex multimodal content. For image reconstruction, we apply Stable Diffusion XL (SDXL) 1.0 \cite{podell2023sdxl} as the restoration module for general domains. SDXL 1.0 is an advanced text-to-image diffusion model that can generate images according to a given prompt. Additionally, for the restoration of chest X-ray modality, we use Cheff \cite{weber2023cascaded}, a cascaded chest X-ray latent diffusion model. By default, we generate 5 candidates for the missing modality during the generation process. Since the chosen LMM has already been pre-trained on the COCO dataset, to ensure a fair comparison, we employ pre-trained CLIP \cite{radford2021learning} vision and text encoders to replace the modality backbones of the comparison methods. We conduct all experiments on the PyTorch 2.4.0 platform, running on Ubuntu 20.04 LTS utilizing 4 GPUs (NVIDIA GeForce RTX 4090 with 24 GB of memory). In our setting, we conduct the missing rate $\eta = \{0.3, 0.5, 0.7\}$ to simulate the missing modality scenario during training.

\begin{table*}[ht]
\setlength{\abovecaptionskip}{0cm} 
\setlength{\belowcaptionskip}{-0.2cm} 
\setlength\tabcolsep{2pt}
\begin{center}
\scalebox{0.94}{
\begin{tabular}{l|lc|c|c|c|c|c|c|c|c|c|c|c|c|c|c|c|c|c}
\toprule
 \multicolumn{2}{c}{} & \multicolumn{9}{c}{COCO-2014 \cite{lin2014microsoft}} & \multicolumn{9}{c}{MM-IMDb \cite{demner2016preparing}} \\
 \cmidrule(lr){3-11} \cmidrule(lr){12-20} 
 \multicolumn{2}{c}{Missing Rate $\eta$}   & \multicolumn{3}{c}{0.3} &  \multicolumn{3}{c}{0.5} & \multicolumn{3}{c}{0.7} & \multicolumn{3}{c}{0.3}  & \multicolumn{3}{c}{0.5} & \multicolumn{3}{c}{0.7}  \\
\cmidrule(lr){3-5} \cmidrule(lr){6-8} \cmidrule(lr){9-11}  \cmidrule(lr){12-14} \cmidrule(lr){15-17} \cmidrule(lr){18-20} 
 \multicolumn{2}{c}{Method} & F1 & AP & SS &  F1  & AP & SS &  F1  & AP & SS &  F1  & AP & SS &  F1  & AP & SS &  F1  & AP & SS  \\
\midrule

\multicolumn{2}{c}{\textcolor{gray}{Baseline (complete)}} & \multicolumn{8}{c}{\textcolor{gray}{F1: 78.3  $|$  AP: 84.6 $|$ SS: -}} & & \multicolumn{8}{c}{\textcolor{gray}{F1: 56.2  $|$  AP: 62.7 $|$ SS: -}} \\
\multicolumn{2}{c}{Baseline (remove missing)}   & 75.8 & 80.8 & - & 73.1 & 79.3 & - & 70.3 & 77.5 & - & 51.8 & 58.5 & - & 50.3 & 56.1 & - & 47.2 & 53.9 & - \\

\midrule

\multicolumn{2}{c}{MPMM \cite{lee2023multimodal} (CVPR'23)}& 76.2 & 82.0 & - & 74.7 & 80.5 & - & 71.0 & 78.8 & - & 53.6 & 59.8 & - & 51.1 & 56.9 & - & 48.5 & 55.7 & - \\
\multicolumn{2}{c}{MPLMM \cite{guo-etal-2024-multimodal} (ACL'24)} & \underline{77.1} & \underline{82.6} & - & \underline{75.2} & \underline{81.3} & - & \underline{72.3} &   \underline{80.1} & - &  \underline{53.9}  & \underline{60.3}  & - &  \underline{52.8} &  \underline{57.3}  & - &  \underline{49.1}  &  \underline{56.2}  & - \\

\multicolumn{2}{c}{ MMIN \cite{zhao2021missing} (ACL'21)}& 73.2 & 78.3 & \underline{37.1} & 71.4 & 77.5 & \underline{36.7} & 70.5 & 76.4 & \underline{36.0} & 50.1 & 53.8 & 26.7 & 49.5 & 51.7 & \underline{26.3} & 44.6 & 50.8 & \underline{24.4} \\
\multicolumn{2}{c}{DiCMoR \cite{wang2023distribution} (CVPR'23)}& 65.3 & 74.4 & 34.3 & 59.7 & 67.1 & 33.5 & 55.3 & 64.0 & 31.9 & 49.2 & 54.7 & \underline{26.9} & 43.7 & 50.8 & 25.9 & 30.5 & 41.7 & 23.1 \\

\midrule

\multicolumn{2}{c}{Ours (Qwen-VL-2B)} & 76.1 & 81.8 & 39.8 & 74.3 & 79.9 & 37.1 & 70.7 & 78.2 & 36.3 & 52.4 & 58.9 & 28.8 & 51.9 & 57.1 & 28.3 & 50.3 & 56.5 & 28.1 \\
\multicolumn{2}{c}{Ours (Qwen-VL-7B)} & \textbf{77.5} & \textbf{82.9} & \textbf{40.4} & \textbf{77.5} & \textbf{82.8} & \textbf{40.2} & \textbf{77.9} & \textbf{83.5} & \textbf{38.2} & \textbf{54.7} & \textbf{60.9} & \textbf{33.5} & \textbf{54.9} & \textbf{61.3} & \textbf{32.7} & \textbf{55.2} & \textbf{61.8} & \textbf{32.3} \\

\midrule
\multicolumn{2}{c}{\textcolor{gray}{$\Delta$ Complete Baseline}} & \color{gray} -2.8 & \color{gray} -1.7 & -  & \color{gray} -0.8 & \color{gray} -1.8 & - & \color{gray} -0.4 & \color{gray} -1.1 & - & \color{gray} -1.5 & \color{gray} -1.8 & - & \color{gray} -1.3 & \color{gray} -1.4 & - & \color{gray} -1.0 & \color{gray} -0.9 & - \\

\multicolumn{2}{c}{$\Delta$ SOTA} & \color{red} +0.4 & \color{red} +0.3 & \color{red} +3.3 & \color{red} +2.3 & \color{red} +1.5 & \color{red} +3.7 & \color{red} +5.6 & \color{red} +3.4 & \color{red} +2.5 & \color{red} +0.8 & \color{red} +0.6 & \color{red} +6.6 & \color{red} +2.1 & \color{red} +4.0 & \color{red} +6.4 & \color{red} +6.1 & \color{red} +5.6 & \color{red} +7.9 \\


\bottomrule
\end{tabular}
}
\end{center}
\caption{\textbf{Quantitative analysis results (\%) on COCO-2014 and MM-IMDb datasets.} \textbf{Bold} denotes the best results and \underline{underline} denotes the second-best. SS (\%) refers to the average similarity score, which is used to assess the generation quality of imputation-based methods. A higher score indicates better quality. `-' indicates that the metric is not applicable. All results are reproduced using the officially released code.}
\label{tab:quantitative-result}       
\end{table*}

\begin{table}[h]
\setlength{\abovecaptionskip}{0cm} 
\setlength{\belowcaptionskip}{-0.2cm} 
\setlength\tabcolsep{2pt}
\begin{center}
\scalebox{0.9}{
\begin{tabular}{lc|c|c|c|c|c}
\toprule

Missing Rate $\eta$ & \multicolumn{3}{c}{0.3} & \multicolumn{3}{c}{0.7}\\
\cmidrule(lr){2-4} \cmidrule(lr){5-7}
Method & F1 & AP & SS & F1 & AP & SS \\
\midrule
\textcolor{gray}{Baseline (complete)} & \multicolumn{6}{c}{\textcolor{gray}{F1: 57.0 $|$ AP: 75.7$|$ SS: -}} \\
Baseline (remove missing) & 49.1 & 71.4 & -  & 31.5 & 56.2 & - \\
\midrule
MPMM \cite{lee2023multimodal} (CVPR'23) & \underline{49.9} & 71.8 & - & \underline{36.8} & 61.4 & - \\
MPLMM \cite{guo-etal-2024-multimodal} (ACL'24) & 49.3 & \underline{72.7} & - & 35.2 & \underline{61.9} & -\\
MMIN \cite{zhao2021missing} (ACL'21) & 37.3 & 64.2 & 17.3 & 26.7 & 50.1 & 10.2  \\
DiCMoR \cite{wang2023distribution} (CVPR'23) & 40.5 & 69.1 & \underline{18.1} & 29.8 & 53.6 & \underline{13.3} \\
\midrule
Ours (Qwen-VL-2B)  & 51.4 & 72.0 & 21.9 & 41.1 & 68.9 & 17.7 \\
Ours (Qwen-VL-7B)  & \textbf{53.6} & \textbf{73.9} & \textbf{22.6} & \textbf{46.3} & \textbf{70.5} & \textbf{19.8} \\
\midrule
\textcolor{gray}{$\Delta$ Complete Baseline} &  \color{gray} -3.4 & \color{gray} -1.8 & \color{gray} - & \color{gray} -10.7 & \color{gray} -5.2 & \color{gray} - \\
$\Delta$ SOTA & \color{red} +3.7 & \color{red} +1.2 & \color{red} +4.5 & \color{red} +9.5 & \color{red} +8.6 & \color{red} +6.5 \\
\bottomrule
\end{tabular}}
\end{center}
\caption{\textbf{Quantitative results (\%) on IU X-ray datasets.} \textbf{Bold} denotes the best results and \underline{underline} denotes the second-best. SS (\%) refers to the average similarity score, which is used to assess the generation quality of imputation-based methods. A higher score indicates better quality. `-' indicates that the metric is not applicable. All results are reproduced using the officially released code.}
\label{tab:classification-result}       
\vspace{-10pt}
\end{table}

\subsection{Quantitative Analysis}

In Tables \ref{tab:quantitative-result} and \ref{tab:classification-result}, we present the results of our method compared to others across different missing data ratios, as measured by F1, AP, and mean similarity score (SS) on three datasets \footnote{More results (more modalities) are presented in the appendix.}. To comprehensively evaluate the performance of various approaches, we introduce two baselines: one trained on complete data and another trained on data with missing entries removed. Additionally, to fair comparison regarding the scale of model parameters, the proposed method based on the Qwen-VL-2B were incorporated.

Results from general domain datasets (COCO-2014 and MM-IMDb) indicate that most MMC methods outperform the baseline where missing data is simply removed. Our method demonstrates superior performance across different missing rates, particularly at a missing rate of 0.7, where it shows significant improvements in both F1 and AP metrics. We attribute these enhancements to the synthetic data generated by our method, which appears to bolster downstream task performance. This phenomenon is also observed in \cite{gu2024infinity, wang2023see}. However, it should be noted that the synthetic data may lead to a slight decline in mean similarity score.

On the other hand, results from the OOD dataset highlight our method's superior domain adaptation capability compared to other methods. Our approach not only excels in classification metrics but also significantly surpasses other imputation-based methods, such as MMIN and DiCMoR, in missing data generation metrics (SS). This advantage can be credited to our method's ability to harness the few-shot learning and in-context learning capabilities of LMM, effectively mitigating the impact of varying degrees of data incompleteness.

\subsection{Ablation Study}

\noindent \textbf{Q1: Does `knowledge' really help MMC?} We study the impact of different components under general and OOD scenarios with the same missing rate.  As shown in Table \ref{tab:ab-study-knowledge}, we report the results of various components. For the generation process, we analyze the variant without knowledge modeling\footnote{We directly use LMM to generate descriptions of available modalities.} (row 1) and the variant with random ranking (row 2). Additionally, in examining the ranking module, we assess the variant with random ranking (row 3), the variant without knowledge graph ranking (row 4), and the variant without using semantic similarity score ranking (row 5). The results demonstrate that the variant without knowledge modeling shows a significant decline across all metrics under OOD scenarios. Furthermore, the semantic similarity ranking strategy in the ranking module is crucial for the effectiveness of the MMC. In general, the study reveals that knowledge modeling is the most critical component of our method.
\begin{table}[h]
\setlength{\abovecaptionskip}{0cm} 
\setlength{\belowcaptionskip}{-0.2cm} 
\setlength\tabcolsep{2pt}
\begin{center}
\scalebox{0.85}{
\begin{tabular}{cllc|c|c|c|c|c}
\toprule
\multicolumn{3}{c}{} & \multicolumn{3}{c}{MM-IMDb} & \multicolumn{3}{c}{IU X-ray} \\

\multicolumn{3}{c}{Missing Rate $\eta$} & \multicolumn{3}{c}{0.7} & \multicolumn{3}{c}{0.7}\\
\cmidrule(lr){4-6} \cmidrule(lr){7-9}
\multicolumn{3}{c}{Variants} & F1 & AP & SS & F1 & AP & SS \\
\midrule
0 & \multicolumn{2}{c}{Baseline (Qwen-VL-7B)} & 55.2 & 61.8 & 32.3 & 46.3 & 70.5 & 19.8 \\

\midrule
1 & \multicolumn{2}{c}{ \textit{w/o} Knowledge Modeling} & -1.3 & -3.6 & -8.8 & \color{red}-17.5 & \color{red}-29.2 & \color{red}-13.7 \\
2 & \multicolumn{2}{c}{+ Random Ranking} & -1.6 & -4.1 & -9.9 & \color{red}-19.3 & \color{red}-31.8 & \color{red}-15.0 \\
\midrule
3 & \multicolumn{2}{c}{Random Ranking} & -0.5 & -2.7 & -0.6 & -3.8 & -7.1 & -4.7 \\
4 & \multicolumn{2}{c}{\textit{w/o} Knowledge Ranking} & -0.2 & -0.8 & -0.1 & -1.9 & -2.7 & -1.1  \\
5 & \multicolumn{2}{c}{\textit{w/o} Semantic Ranking} & -0.2 & -1.0 & -0.3 & -2.4 & -3.3 & -1.6 \\
\bottomrule
\end{tabular}}
\end{center}
\caption{\textbf{The impact of various components.} We report the comparison results between different combinations and the baseline.}
\label{tab:ab-study-knowledge}       
\end{table}

\begin{figure}[h]
    \centering
    \begin{subfigure}[b]{0.41\textwidth}
        \centering
        \includegraphics[width=\textwidth]{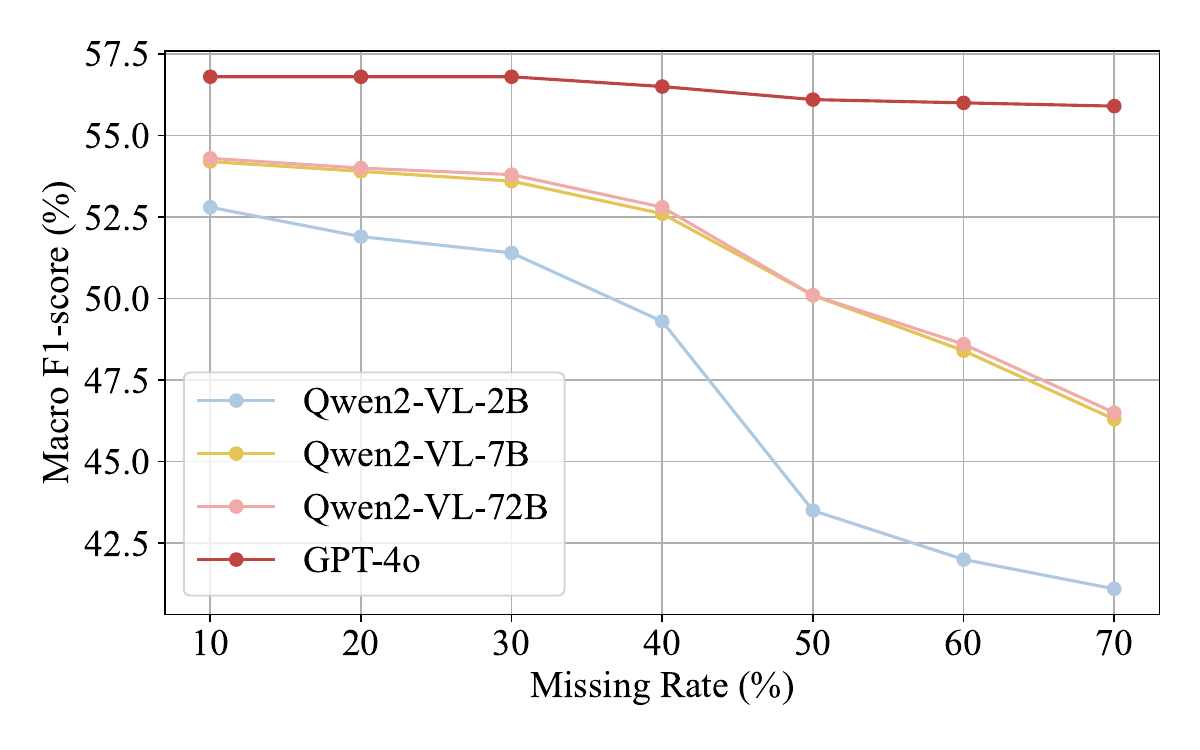}
        \caption{F1 score at different missing rates.}
        \label{fig:ab-model-scale-a}
    \end{subfigure}
    \begin{subfigure}[b]{0.4\textwidth}
        \centering
        \includegraphics[width=\textwidth]{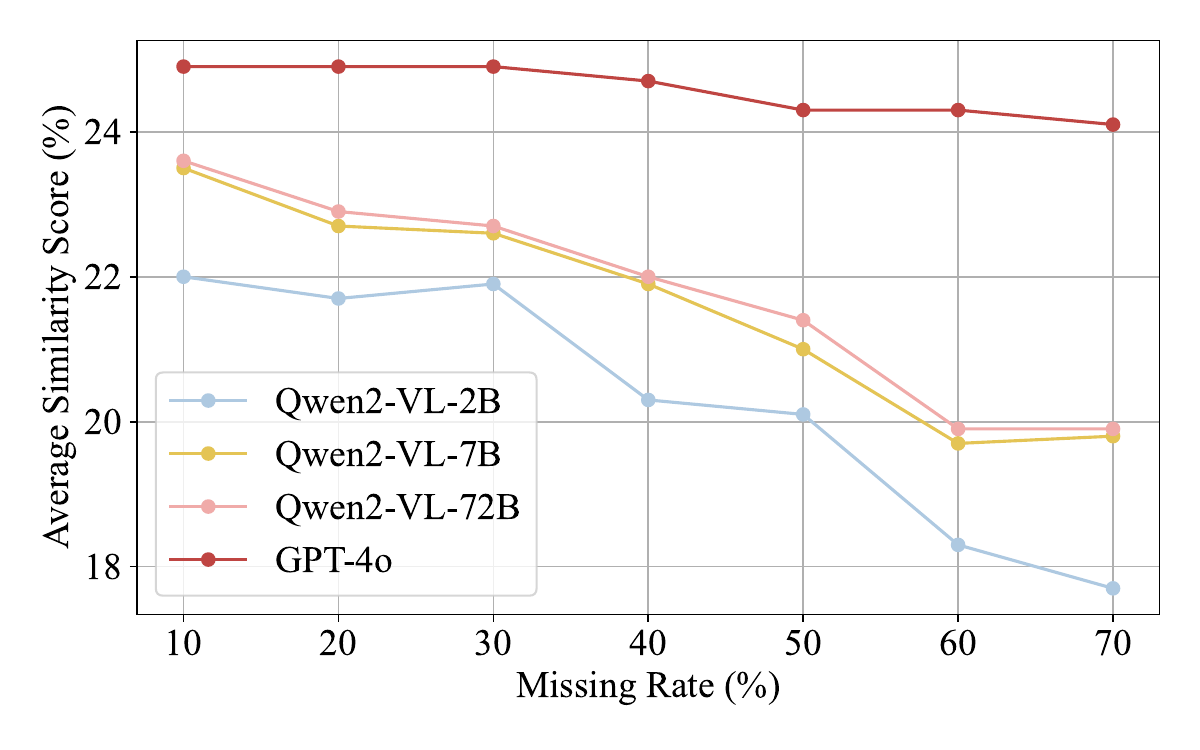}
        \caption{Average similarity score at different missing rates. }
        \label{fig:ab-model-scale-b}
    \end{subfigure}
    \caption{The impact of different model's parameters scales and missing rates.}
    \label{fig:ab-model-scale}
\vspace{-10pt}
\end{figure}

\noindent \textbf{Q2: Does the scale of model parameters affect MMC?} We evaluate the performance of our approach using different model scales under at various missing rates. Specifically, we conduct the 2B, 7B, and 72B variants of the open-source large model Qwen-VL \cite{wang2024qwen2}, as well as OpenAI's GPT-4o \cite{hurst2024gpt} on the IU X-ray dataset. For a fair comparison, we leverage these models solely for handling the knowledge modeling and integration in steps 1 and 2 of our method, keeping the modality generators unchanged. As illustrated in Fig. \ref{fig:ab-model-scale}, GPT-4o demonstrates significant superiority and robustness. These results demonstrate that as the scale of model parameters increases, the quality of knowledge modeling in our approach also improves.


\noindent \textbf{Q3: Does generating more missing data improve performance?} We evaluate the impact of different numbers of generation candidates in step 2 of our method on the MM-IMDb and IU X-ray datasets. Specifically, we conduct the settings with generation candidates \( n = \{1, 5, 10, 15\} \), and the results are presented in Table \ref{tab:ab-study-generation}. The findings indicate that generating only one candidate significantly reduces the missing generation quality. While using a higher number of generation candidates can lead to slight improvements across all metrics, it also substantially increases inference time. Therefore, setting \( n = 5 \) provides a balance between inference time and performance.

\begin{table}[t]
\setlength{\abovecaptionskip}{0cm} 
\setlength{\belowcaptionskip}{-0.2cm} 
\setlength\tabcolsep{2pt}
\begin{center}
\scalebox{0.9}{
\begin{tabular}{lcc|c|c|c|c|c}
\toprule

 & & \multicolumn{3}{c}{MM-IMDb} & \multicolumn{3}{c}{IU X-ray} \\

Missing Rate $\eta$ &  & \multicolumn{3}{c}{0.7} & \multicolumn{3}{c}{0.7}\\
\cmidrule(lr){3-5} \cmidrule(lr){6-8}
Candidates $n$ & time & F1 & AP & SS & F1 & AP & SS \\
\midrule
\color{gray}$n = 5$ (default) & 40s &  \color{gray} 55.2 & \color{gray} 61.8 & \color{gray} 32.3 & \color{gray} 46.3 & \color{gray} 70.5 & \color{gray} 19.8 \\
\midrule
$n = 1$ & 7s &  -0.7 & -2.8 & -7.3 & -8.5 & -16.4 & -14.7 \\
$n = 10$ & 83s &  +0.4 & +0.9 & +0.1 & +0.0 & +0.1 & +0.0 \\
$n = 15$ & 132s & +0.4 & +0.9 & +0.1 & +0.1 & +1.3 & +0.3 \\
\bottomrule
\end{tabular}}
\end{center}
\caption{The impact of different generation candidates.}
\label{tab:ab-study-generation}       
\vspace{-10pt}
\end{table}








\subsection{Visualization Analysis}
\begin{figure*}[ht]
  \centering
  \includegraphics[width=\linewidth]{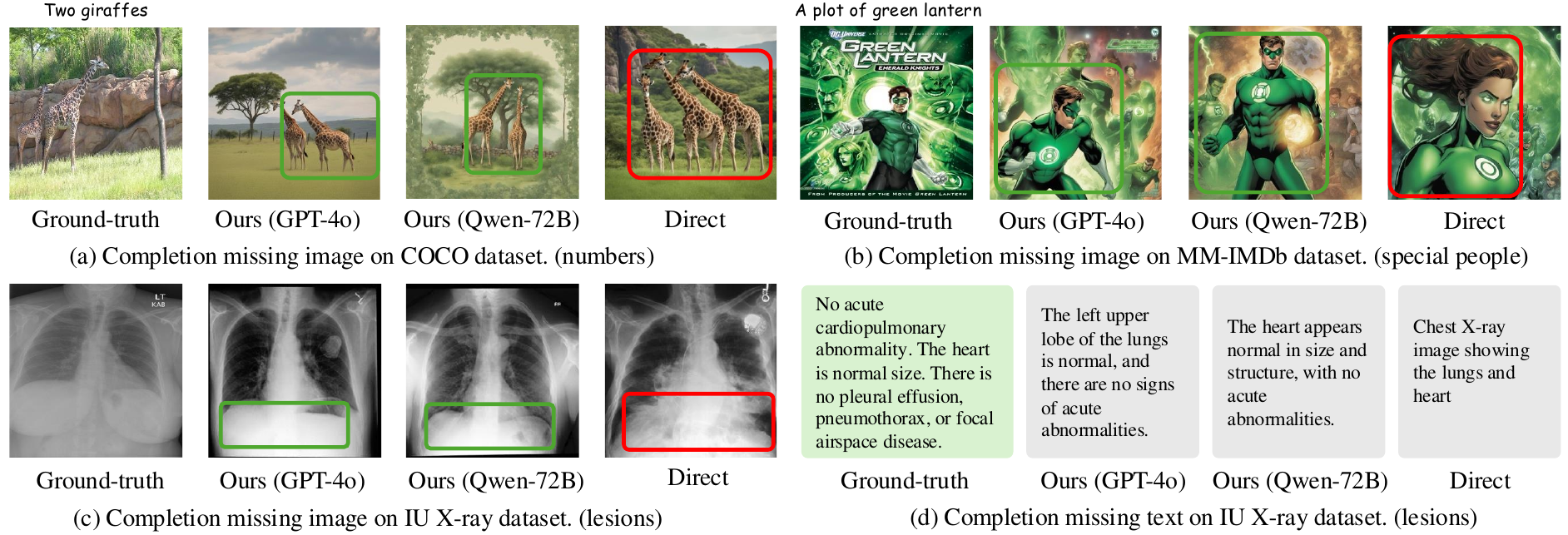}
  \caption{\textbf{Visualization analysis.} We present the results of multi-modality completion across different datasets. These results include visualizations related to the number of objects and specific people cases in general domain, as well as lesion completion in the medical domain. The green box represents the part that is close to the ground-truth, and the red box represents the wrong part.}
  \label{fig:vis-1}
\vspace{-10pt}
\end{figure*}

To better understanding the differences between our approach and conditional generation, we present completion results produced by our method, as shown in Fig. \ref{fig:vis-1}. We provide visualization results across different missing modalities for three datasets\footnote{More visualization are presented in appendix.}. In the general domain, our knowledge modeling module focuses on understanding the quantity of objects, their attributes, and the contextual environment. The results in Fig. \ref{fig:vis-1}a and Fig. \ref{fig:vis-1}b indicate that our method is more similar with the original missing modality compared to direct generation approaches. In the medical domain, incorporating knowledge of different lesions enables the LMM to understand the relationships between various regions in chest X-rays and the content described by the modality. Fig. \ref{fig:vis-1}c and Fig. \ref{fig:vis-1}d show that our method provides a more reliable strategy for missing data completion than direct generation.

\section{Related Work}
\label{sec:related-work}

\subsection{Missing Multimodal Learning}

Missing multimodal learning enables models to train and inference effectively even when some modalities are absent. The missing multimodal learning methods can be categorized into imputation-based and non-imputation-based approaches. Imputation-based methods predict or recover missing data by learning interactions between modalities. Typical strategies include filling with special values such as zero or the average value, which are unsuitable for high-dimensional data \cite{parthasarathy2020training} . Advanced methods like SMIL \cite{ma2021smil} use meta-learning to simulate missing modalities. Wang \textit{et al.} \cite{wang2023multi} propose a multitask framework for missing modality learning, which learns shared features across different tasks and uses these shared features to reconstruct missing modalities. Lian \textit{et al.} \cite{lian2023gcnet} use graph neural networks to explore relationships between modalities and recover missing ones, while Zhao \textit{et al.} \cite{zhao2021missing} and Pham \textit{et al.} \cite{pham2019found} leverage cycle consistency to capture cross-modal interactions for predicting missing modalities. 


On the other hand, non-imputation-based methods focus on fusion strategies or missing indicators instead of predicting missing data \cite{ke2023clustering}. Wang \textit{et al.} \cite{wang2020transmodality} and Han \textit{et al.} \cite{han2019implicit} use translational semantics for implicit fusion. Ma \textit{et al.} \cite{ma2022multimodal} leverage transformers' variable input capability. Lee \textit{et al.} \cite{lee2023multimodal} and Guo \textit{et al.} \cite{guo-etal-2024-multimodal} introduce learnable missing semantic tokens, while Zeng \textit{et al.} \cite{zeng2022tag} use fixed tags to identify missing inputs. Current methods often require pre-training with complete data and show limited domain transfer. Our approach leverages LMM to predict missing modalities using domain knowledge without extra pre-training, employing few-shot learning for flexible adaptation.

\subsection{Large Multimodal Models}

Large multimodal models (LMMs) integrate diverse data types, such as text, images, and audio, to enhance understanding and interaction capabilities. Early works focused on fusion techniques to combine these modalities, like early \cite{chen2020uniter}, late \cite{evangelopoulos2013multimodal}, or hybrid fusion \cite{prabu2019multimodal}. Recent advancements \cite{team2024chameleon, team2024gemini}, such as OpenAI's GPT-4o \cite{hurst2024gpt} and Google's PaLM-E \cite{driess2023palm}, have achieved remarkable success by increasing the scale of both pre-training data and model parameters. These private ultra-models (greater than 100 billion parameters) exhibit robust generalization across various domains, which enables applications in image captioning, visual question answering, and multimodal dialogue. On the other hand, some open-source, medium- to small-scale (less than 100 billion parameters) models, such as LLaVA \cite{liu2024visual}, Qwen-VL \cite{wang2024qwen2}, and InternVL \cite{chen2024internvl}, have matched the performance of these ultra-large models. The common feature among them is their rich prior knowledge, understanding, and OOD capabilities, which form the foundation of training-free missing multi-modality completion.

\subsection{Conditional Generation}

Conditional generation creates new data samples based on specific inputs, enabling models to generate tailored outputs. Early foundational work in this area includes Generative Adversarial Networks (GANs) \cite{goodfellow2014generative} and Variational Autoencoders (VAEs) \cite{kingma2013auto}, with notable advancements such as conditional GANs \cite{mirza2014conditional} and Conditional VAEs (CVAEs) \cite{sohn2015learning}. Recent progress has introduced diffusion models \cite{song2020denoising, li2022diffusion, ruiz2023dreambooth}, which have demonstrated the ability to generate more realistic content, as exemplified by systems like DALL-E \cite{ramesh2021zero} and Stable Diffusion \cite{podell2023sdxl}. ControlNet \cite{zhang2023adding} further improves user control. The key distinction between these approaches and our method lies in their respective objectives. Unlike these approaches, which focus on diverse content generation, our method emphasizes understanding modality interactions to \textbf{accurately} complete \textbf{missing content}. 


\section{Discussion and Limitations}

\noindent \textbf{Why we need the training-free MMC?} Previous approaches for handling missing modalities have demonstrated limited generalizability across different domains and constrained data flexibility. For example, recent prompt-based methods such as MPMM \cite{lee2023multimodal} and MPLMM \cite{guo-etal-2024-multimodal} have shown significantly lower performance in general domains compared to their effectiveness in the medical domain, as reflected in our experimental results. Additionally, these methods depend on pre-trained models that are highly tailored to the target domain. In data-hungry scenarios, such as the medical field where data often contain missing elements, these constraints greatly impact the reliability of model decisions. In recent years, LMMs have exhibited remarkable zero-shot capabilities across various domains. Consequently, leveraging these models in a training-free manner to address the MMC problem presents a promising and cost-effective research direction. Our experimental results corroborate this perspective, and we hope these findings will encourage the community to further explore the potential of LMMs for addressing MMC challenges.

\noindent \textbf{Why `knowledge bridger' can help MMC?} We believe that the key to addressing the MMC problem using LMM lies in accurately generating missing data and effectively ranking the generation candidates. Our experimental results show that directly employing LMM to generate the missing modality does not ensure the accuracy required for MMC. Our proposed approach first involves mining internal knowledge from the available modalities, enabling LMM to understand intra-modal interactions. This structured knowledge is then used to address challenges in both generation and ranking. However, using LMM inherently leads to hallucinations. In future work, we plan to employ more robust knowledge extraction methods, such as retrieval-augmented generation \cite{guo2024lightrag}, to mitigate hallucinations\footnote{Potential hallucinations of our method are presented in appendix.} and further enhance the accuracy of missing data generation.

\noindent \textbf{Limitations.} Our method focuses exclusively on image and text modalities, leaving its performance on other modalities, such as speech and depth, yet to be explored. The approach emphasizes the automatic extraction of inter-modal knowledge and the completion of missing modalities through domain knowledge. Thus, in the future, adaptation to other modalities is possible by defining a more comprehensive modality knowledge. Some promising works \cite{girdhar2023imagebind, lyu2024unibind} show that one modality, such as image or text, can be connected to any other modality. Additionally, we observe that while our method enhances classification performance under a high missing rate (e.g., 0.7), it paradoxically results in a decrease in the similarity scores of the completed modalities. Therefore, there remains substantial potential for further exploration to develop more robust generation and ranking strategies in the future.

\section{Conclusion}
In this paper, we propose a novel MMC challenge and introduce a training-free MMC pipeline called `Knowledge Bridger' based on LMM. Our findings demonstrate that LMM can autonomously mine knowledge across modalities and leverage this structured knowledge to enhance the accuracy of missing modality generation and ranking. Furthermore, experimental results indicate that as the model's parameter scale increases, our method shows improved knowledge extraction capabilities.

\textbf{Acknowledgment.} This work is supported by the Guangdong Natural Science Funds for Distinguished Young Scholars (Grant 2023B1515020097), the AI Singapore Programme under the National Research Foundation Singapore (Grant AISG3-GV-2023-011), the Beijing Natural Science Foundation (No.4234086), the Natural Science Foundation of China (No. 62192782), and the Lee Kong Chian Fellowships.
{
    \small
    \bibliographystyle{ieeenat_fullname}
    \bibliography{main}

\begin{thebibliography}{57}
\providecommand{\natexlab}[1]{#1}
\providecommand{\url}[1]{\texttt{#1}}
\expandafter\ifx\csname urlstyle\endcsname\relax
  \providecommand{\doi}[1]{doi: #1}\else
  \providecommand{\doi}{doi: \begingroup \urlstyle{rm}\Url}\fi

\bibitem[Arevalo et~al.(2017)Arevalo, Solorio, Montes-y G{\'o}mez, and Gonz{\'a}lez]{arevalo2017gated}
John Arevalo, Thamar Solorio, Manuel Montes-y G{\'o}mez, and Fabio~A Gonz{\'a}lez.
\newblock Gated multimodal units for information fusion.
\newblock In \emph{ICLR}, 2017.

\bibitem[Bianco et~al.(2023)Bianco, Celona, Donzella, and Napoletano]{bianco2023improving}
Simone Bianco, Luigi Celona, Marco Donzella, and Paolo Napoletano.
\newblock Improving image captioning descriptiveness by ranking and llm-based fusion.
\newblock \emph{arXiv preprint arXiv:2306.11593}, 2023.

\bibitem[Cai et~al.(2018)Cai, Wang, Gao, Shen, and Ji]{cai2018deep}
Lei Cai, Zhengyang Wang, Hongyang Gao, Dinggang Shen, and Shuiwang Ji.
\newblock Deep adversarial learning for multi-modality missing data completion.
\newblock In \emph{SIGKDD}, pages 1158--1166, 2018.

\bibitem[Chen et~al.(2020)Chen, Li, Yu, El~Kholy, Ahmed, Gan, Cheng, and Liu]{chen2020uniter}
Yen-Chun Chen, Linjie Li, Licheng Yu, Ahmed El~Kholy, Faisal Ahmed, Zhe Gan, Yu Cheng, and Jingjing Liu.
\newblock Uniter: Universal image-text representation learning.
\newblock In \emph{ECCV}, pages 104--120. Springer, 2020.

\bibitem[Chen et~al.(2024)Chen, Wu, Wang, Su, Chen, Xing, Zhong, Zhang, Zhu, Lu, et~al.]{chen2024internvl}
Zhe Chen, Jiannan Wu, Wenhai Wang, Weijie Su, Guo Chen, Sen Xing, Muyan Zhong, Qinglong Zhang, Xizhou Zhu, Lewei Lu, et~al.
\newblock Internvl: Scaling up vision foundation models and aligning for generic visual-linguistic tasks.
\newblock In \emph{CVPR}, pages 24185--24198, 2024.

\bibitem[Cohen~Kalafut et~al.(2023)Cohen~Kalafut, Huang, and Wang]{cohen2023joint}
Noah Cohen~Kalafut, Xiang Huang, and Daifeng Wang.
\newblock Joint variational autoencoders for multimodal imputation and embedding.
\newblock \emph{Nature Machine Intelligence}, 5\penalty0 (6):\penalty0 631--642, 2023.

\bibitem[Demner-Fushman et~al.(2016)Demner-Fushman, Kohli, Rosenman, Shooshan, Rodriguez, Antani, Thoma, and McDonald]{demner2016preparing}
Dina Demner-Fushman, Marc~D Kohli, Marc~B Rosenman, Sonya~E Shooshan, Laritza Rodriguez, Sameer Antani, George~R Thoma, and Clement~J McDonald.
\newblock Preparing a collection of radiology examinations for distribution and retrieval.
\newblock \emph{Journal of the American Medical Informatics Association}, 23\penalty0 (2):\penalty0 304--310, 2016.

\bibitem[Driess et~al.(2023)Driess, Xia, Sajjadi, Lynch, Chowdhery, Ichter, Wahid, Tompson, Vuong, Yu, et~al.]{driess2023palm}
Danny Driess, Fei Xia, Mehdi~SM Sajjadi, Corey Lynch, Aakanksha Chowdhery, Brian Ichter, Ayzaan Wahid, Jonathan Tompson, Quan Vuong, Tianhe Yu, et~al.
\newblock Palm-e: An embodied multimodal language model.
\newblock In \emph{ICML}, 2023.

\bibitem[Evangelopoulos et~al.(2013)Evangelopoulos, Zlatintsi, Potamianos, Maragos, Rapantzikos, Skoumas, and Avrithis]{evangelopoulos2013multimodal}
Georgios Evangelopoulos, Athanasia Zlatintsi, Alexandros Potamianos, Petros Maragos, Konstantinos Rapantzikos, Georgios Skoumas, and Yannis Avrithis.
\newblock Multimodal saliency and fusion for movie summarization based on aural, visual, and textual attention.
\newblock \emph{IEEE TMM}, 15\penalty0 (7):\penalty0 1553--1568, 2013.

\bibitem[Girdhar et~al.(2023)Girdhar, El-Nouby, Liu, Singh, Alwala, Joulin, and Misra]{girdhar2023imagebind}
Rohit Girdhar, Alaaeldin El-Nouby, Zhuang Liu, Mannat Singh, Kalyan~Vasudev Alwala, Armand Joulin, and Ishan Misra.
\newblock Imagebind: One embedding space to bind them all.
\newblock In \emph{CVPR}, pages 15180--15190, 2023.

\bibitem[Goodfellow et~al.(2014)Goodfellow, Pouget-Abadie, Mirza, Xu, Warde-Farley, Ozair, Courville, and Bengio]{goodfellow2014generative}
Ian Goodfellow, Jean Pouget-Abadie, Mehdi Mirza, Bing Xu, David Warde-Farley, Sherjil Ozair, Aaron Courville, and Yoshua Bengio.
\newblock Generative adversarial nets.
\newblock In \emph{NIPS}, 2014.

\bibitem[Gu et~al.(2024)Gu, Zhang, Zhou, Yu, Xing, Wang, Cao, Jia, Zhang, Wang, et~al.]{gu2024infinity}
Shuhao Gu, Jialing Zhang, Siyuan Zhou, Kevin Yu, Zhaohu Xing, Liangdong Wang, Zhou Cao, Jintao Jia, Zhuoyi Zhang, Yixuan Wang, et~al.
\newblock Infinity-mm: Scaling multimodal performance with large-scale and high-quality instruction data.
\newblock \emph{arXiv preprint arXiv:2410.18558}, 2024.

\bibitem[Guo et~al.(2024{\natexlab{a}})Guo, Jin, and Zhao]{guo-etal-2024-multimodal}
Zirun Guo, Tao Jin, and Zhou Zhao.
\newblock Multimodal prompt learning with missing modalities for sentiment analysis and emotion recognition.
\newblock In \emph{ACL}, pages 1726--1736, 2024{\natexlab{a}}.

\bibitem[Guo et~al.(2024{\natexlab{b}})Guo, Xia, Yu, Ao, and Huang]{guo2024lightrag}
Zirui Guo, Lianghao Xia, Yanhua Yu, Tu Ao, and Chao Huang.
\newblock Lightrag: Simple and fast retrieval-augmented generation.
\newblock \emph{arXiv preprint arXiv:2410.05779}, 2024{\natexlab{b}}.

\bibitem[Han et~al.(2019)Han, Zhang, Ren, and Schuller]{han2019implicit}
Jing Han, Zixing Zhang, Zhao Ren, and Bj{\"o}rn Schuller.
\newblock Implicit fusion by joint audiovisual training for emotion recognition in mono modality.
\newblock In \emph{ICASSP}, pages 5861--5865. IEEE, 2019.

\bibitem[Hurst et~al.(2024)Hurst, Lerer, Goucher, Perelman, Ramesh, Clark, Ostrow, Welihinda, Hayes, Radford, et~al.]{hurst2024gpt}
Aaron Hurst, Adam Lerer, Adam~P Goucher, Adam Perelman, Aditya Ramesh, Aidan Clark, AJ Ostrow, Akila Welihinda, Alan Hayes, Alec Radford, et~al.
\newblock Gpt-4o system card.
\newblock \emph{arXiv preprint arXiv:2410.21276}, 2024.

\bibitem[Jiang et~al.(2024)Jiang, Tang, Gao, Du, He, and Li]{jiang2024delving}
Xin Jiang, Hao Tang, Junyao Gao, Xiaoyu Du, Shengfeng He, and Zechao Li.
\newblock Delving into multimodal prompting for fine-grained visual classification.
\newblock In \emph{AAAI}, pages 2570--2578, 2024.

\bibitem[Ke et~al.(2023)Ke, Chao, Wang, Xu, Zhu, and Yu]{ke2023clustering}
Guanzhou Ke, Guoqing Chao, Xiaoli Wang, Chenyang Xu, Yongqi Zhu, and Yang Yu.
\newblock A clustering-guided contrastive fusion for multi-view representation learning.
\newblock \emph{IEEE T-CSVT}, 34\penalty0 (4):\penalty0 2056--2069, 2023.

\bibitem[Kingma(2013)]{kingma2013auto}
Diederik~P Kingma.
\newblock Auto-encoding variational bayes.
\newblock \emph{arXiv preprint arXiv:1312.6114}, 2013.

\bibitem[Kojima et~al.(2022)Kojima, Gu, Reid, Matsuo, and Iwasawa]{kojima2022large}
Takeshi Kojima, Shixiang~Shane Gu, Machel Reid, Yutaka Matsuo, and Yusuke Iwasawa.
\newblock Large language models are zero-shot reasoners.
\newblock \emph{NIPS}, 35:\penalty0 22199--22213, 2022.

\bibitem[Lee et~al.(2023)Lee, Tsai, Chiu, and Lee]{lee2023multimodal}
Yi-Lun Lee, Yi-Hsuan Tsai, Wei-Chen Chiu, and Chen-Yu Lee.
\newblock Multimodal prompting with missing modalities for visual recognition.
\newblock In \emph{CVPR}, pages 14943--14952, 2023.

\bibitem[Li et~al.(2023)Li, Li, Savarese, and Hoi]{li2023blip}
Junnan Li, Dongxu Li, Silvio Savarese, and Steven Hoi.
\newblock Blip-2: Bootstrapping language-image pre-training with frozen image encoders and large language models.
\newblock In \emph{ICML}, pages 19730--19742. PMLR, 2023.

\bibitem[Li et~al.(2022)Li, Thickstun, Gulrajani, Liang, and Hashimoto]{li2022diffusion}
Xiang Li, John Thickstun, Ishaan Gulrajani, Percy~S Liang, and Tatsunori~B Hashimoto.
\newblock Diffusion-lm improves controllable text generation.
\newblock In \emph{NIPS}, pages 4328--4343, 2022.

\bibitem[Lian et~al.(2023)Lian, Chen, Sun, Liu, and Tao]{lian2023gcnet}
Zheng Lian, Lan Chen, Licai Sun, Bin Liu, and Jianhua Tao.
\newblock Gcnet: Graph completion network for incomplete multimodal learning in conversation.
\newblock \emph{IEEE T-PAMI}, 45\penalty0 (7):\penalty0 8419--8432, 2023.

\bibitem[Lin et~al.(2014)Lin, Maire, Belongie, Hays, Perona, Ramanan, Doll{\'a}r, and Zitnick]{lin2014microsoft}
Tsung-Yi Lin, Michael Maire, Serge Belongie, James Hays, Pietro Perona, Deva Ramanan, Piotr Doll{\'a}r, and C~Lawrence Zitnick.
\newblock Microsoft coco: Common objects in context.
\newblock In \emph{ECCV}, pages 740--755. Springer, 2014.

\bibitem[Liu et~al.(2024)Liu, Li, Wu, and Lee]{liu2024visual}
Haotian Liu, Chunyuan Li, Qingyang Wu, and Yong~Jae Lee.
\newblock Visual instruction tuning.
\newblock \emph{NIPS}, 36, 2024.

\bibitem[Liu et~al.(2025)Liu, Lv, Kang, Zhang, Liang, and He]{liu2025unsupervised}
Shanglin Liu, Jianming Lv, Jingdan Kang, Huaidong Zhang, Zequan Liang, and Shengfeng He.
\newblock Modfinity: Unsupervised domain adaptation with multimodal information flow intertwining.
\newblock In \emph{CVPR}, 2025.

\bibitem[Lv et~al.(2021)Lv, Liu, and He]{lv2021differentiated}
Jianming Lv, Kaijie Liu, and Shengfeng He.
\newblock Differentiated learning for multi-modal domain adaptation.
\newblock In \emph{ACM MM}, pages 1322--1330, 2021.

\bibitem[Lyu et~al.(2024)Lyu, Zheng, Zhou, and Wang]{lyu2024unibind}
Yuanhuiyi Lyu, Xu Zheng, Jiazhou Zhou, and Lin Wang.
\newblock Unibind: Llm-augmented unified and balanced representation space to bind them all.
\newblock In \emph{CVPR}, pages 26752--26762, 2024.

\bibitem[Ma et~al.(2021)Ma, Ren, Zhao, Tulyakov, Wu, and Peng]{ma2021smil}
Mengmeng Ma, Jian Ren, Long Zhao, Sergey Tulyakov, Cathy Wu, and Xi Peng.
\newblock Smil: Multimodal learning with severely missing modality.
\newblock In \emph{AAAI}, pages 2302--2310, 2021.

\bibitem[Ma et~al.(2022)Ma, Ren, Zhao, Testuggine, and Peng]{ma2022multimodal}
Mengmeng Ma, Jian Ren, Long Zhao, Davide Testuggine, and Xi Peng.
\newblock Are multimodal transformers robust to missing modality?
\newblock In \emph{CVPR}, pages 18177--18186, 2022.

\bibitem[Mirza(2014)]{mirza2014conditional}
Mehdi Mirza.
\newblock Conditional generative adversarial nets.
\newblock \emph{arXiv preprint arXiv:1411.1784}, 2014.

\bibitem[Ngiam et~al.(2011)Ngiam, Khosla, Kim, Nam, Lee, and Ng]{ngiam2011multimodal}
Jiquan Ngiam, Aditya Khosla, Mingyu Kim, Juhan Nam, Honglak Lee, and Andrew~Y Ng.
\newblock Multimodal deep learning.
\newblock In \emph{ICML}, pages 689--696, 2011.

\bibitem[Parthasarathy and Sundaram(2020)]{parthasarathy2020training}
Srinivas Parthasarathy and Shiva Sundaram.
\newblock Training strategies to handle missing modalities for audio-visual expression recognition.
\newblock In \emph{ICMI}, pages 400--404, 2020.

\bibitem[Pham et~al.(2019)Pham, Liang, Manzini, Morency, and P{\'o}czos]{pham2019found}
Hai Pham, Paul~Pu Liang, Thomas Manzini, Louis-Philippe Morency, and Barnab{\'a}s P{\'o}czos.
\newblock Found in translation: Learning robust joint representations by cyclic translations between modalities.
\newblock In \emph{AAAI}, pages 6892--6899, 2019.

\bibitem[Podell et~al.(2024)Podell, English, Lacey, Blattmann, Dockhorn, M{\"u}ller, Penna, and Rombach]{podell2023sdxl}
Dustin Podell, Zion English, Kyle Lacey, Andreas Blattmann, Tim Dockhorn, Jonas M{\"u}ller, Joe Penna, and Robin Rombach.
\newblock Sdxl: Improving latent diffusion models for high-resolution image synthesis.
\newblock In \emph{ICLR}, 2024.

\bibitem[Prabu et~al.(2019)Prabu, Lakshmanan, and Mohammed]{prabu2019multimodal}
S Prabu, M Lakshmanan, and V~Noor Mohammed.
\newblock A multimodal authentication for biometric recognition system using intelligent hybrid fusion techniques.
\newblock \emph{Journal of medical systems}, 43\penalty0 (8):\penalty0 249, 2019.

\bibitem[Radford et~al.(2021)Radford, Kim, Hallacy, Ramesh, Goh, Agarwal, Sastry, Askell, Mishkin, Clark, et~al.]{radford2021learning}
Alec Radford, Jong~Wook Kim, Chris Hallacy, Aditya Ramesh, Gabriel Goh, Sandhini Agarwal, Girish Sastry, Amanda Askell, Pamela Mishkin, Jack Clark, et~al.
\newblock Learning transferable visual models from natural language supervision.
\newblock In \emph{ICML}, pages 8748--8763. PMLR, 2021.

\bibitem[Ramesh et~al.(2021)Ramesh, Pavlov, Goh, Gray, Voss, Radford, Chen, and Sutskever]{ramesh2021zero}
Aditya Ramesh, Mikhail Pavlov, Gabriel Goh, Scott Gray, Chelsea Voss, Alec Radford, Mark Chen, and Ilya Sutskever.
\newblock Zero-shot text-to-image generation.
\newblock In \emph{ICML}, pages 8821--8831. Pmlr, 2021.

\bibitem[Ren et~al.(2021)Ren, Du, Lv, Han, and He]{ren2021learning}
Sucheng Ren, Yong Du, Jianming Lv, Guoqiang Han, and Shengfeng He.
\newblock Learning from the master: Distilling cross-modal advanced knowledge for lip reading.
\newblock In \emph{CVPR}, pages 13325--13333, 2021.

\bibitem[Ruiz et~al.(2023)Ruiz, Li, Jampani, Pritch, Rubinstein, and Aberman]{ruiz2023dreambooth}
Nataniel Ruiz, Yuanzhen Li, Varun Jampani, Yael Pritch, Michael Rubinstein, and Kfir Aberman.
\newblock Dreambooth: Fine tuning text-to-image diffusion models for subject-driven generation.
\newblock In \emph{CVPR}, pages 22500--22510, 2023.

\bibitem[Sohn et~al.(2015)Sohn, Lee, and Yan]{sohn2015learning}
Kihyuk Sohn, Honglak Lee, and Xinchen Yan.
\newblock Learning structured output representation using deep conditional generative models.
\newblock \emph{NIPS}, 28, 2015.

\bibitem[Song et~al.(2020)Song, Meng, and Ermon]{song2020denoising}
Jiaming Song, Chenlin Meng, and Stefano Ermon.
\newblock Denoising diffusion implicit models.
\newblock In \emph{ICLR}, 2020.

\bibitem[Team(2024{\natexlab{a}})]{team2024chameleon}
Chameleon Team.
\newblock Chameleon: Mixed-modal early-fusion foundation models.
\newblock \emph{arXiv preprint arXiv:2405.09818}, 2024{\natexlab{a}}.

\bibitem[Team(2024{\natexlab{b}})]{team2024gemini}
Gemini Team.
\newblock Gemini 1.5: Unlocking multimodal understanding across millions of tokens of context.
\newblock \emph{arXiv preprint arXiv:2403.05530}, 2024{\natexlab{b}}.

\bibitem[Wang et~al.(2023{\natexlab{a}})Wang, Chen, Ma, Avery, Hull, and Carneiro]{wang2023multi}
Hu Wang, Yuanhong Chen, Congbo Ma, Jodie Avery, Louise Hull, and Gustavo Carneiro.
\newblock Multi-modal learning with missing modality via shared-specific feature modelling.
\newblock In \emph{CVPR}, pages 15878--15887, 2023{\natexlab{a}}.

\bibitem[Wang et~al.(2023{\natexlab{b}})Wang, Meng, Weng, He, Wu, and Jiang]{wang2023see}
Junke Wang, Lingchen Meng, Zejia Weng, Bo He, Zuxuan Wu, and Yu-Gang Jiang.
\newblock To see is to believe: Prompting gpt-4v for better visual instruction tuning.
\newblock \emph{arXiv preprint arXiv:2311.07574}, 2023{\natexlab{b}}.

\bibitem[Wang et~al.(2024)Wang, Bai, Tan, Wang, Fan, Bai, Chen, Liu, Wang, Ge, et~al.]{wang2024qwen2}
Peng Wang, Shuai Bai, Sinan Tan, Shijie Wang, Zhihao Fan, Jinze Bai, Keqin Chen, Xuejing Liu, Jialin Wang, Wenbin Ge, et~al.
\newblock Qwen2-vl: Enhancing vision-language model's perception of the world at any resolution.
\newblock \emph{arXiv preprint arXiv:2409.12191}, 2024.

\bibitem[Wang et~al.(2023{\natexlab{c}})Wang, Cui, and Li]{wang2023distribution}
Yuanzhi Wang, Zhen Cui, and Yong Li.
\newblock Distribution-consistent modal recovering for incomplete multimodal learning.
\newblock In \emph{ICCV}, pages 22025--22034, 2023{\natexlab{c}}.

\bibitem[Wang et~al.(2020)Wang, Wan, and Wan]{wang2020transmodality}
Zilong Wang, Zhaohong Wan, and Xiaojun Wan.
\newblock Transmodality: An end2end fusion method with transformer for multimodal sentiment analysis.
\newblock In \emph{WWW}, pages 2514--2520, 2020.

\bibitem[Weber et~al.(2023)Weber, Ingrisch, Bischl, and R{\"u}gamer]{weber2023cascaded}
Tobias Weber, Michael Ingrisch, Bernd Bischl, and David R{\"u}gamer.
\newblock Cascaded latent diffusion models for high-resolution chest x-ray synthesis.
\newblock In \emph{Pacific-Asia Conference on Knowledge Discovery and Data Mining}, pages 180--191. Springer, 2023.

\bibitem[Wei et~al.(2022{\natexlab{a}})Wei, Tay, Bommasani, Raffel, Zoph, Borgeaud, Yogatama, Bosma, Zhou, Metzler, et~al.]{wei2022emergent}
Jason Wei, Yi Tay, Rishi Bommasani, Colin Raffel, Barret Zoph, Sebastian Borgeaud, Dani Yogatama, Maarten Bosma, Denny Zhou, Donald Metzler, et~al.
\newblock Emergent abilities of large language models.
\newblock \emph{arXiv preprint arXiv:2206.07682}, 2022{\natexlab{a}}.

\bibitem[Wei et~al.(2022{\natexlab{b}})Wei, Wang, Schuurmans, Bosma, Xia, Chi, Le, Zhou, et~al.]{wei2022chain}
Jason Wei, Xuezhi Wang, Dale Schuurmans, Maarten Bosma, Fei Xia, Ed Chi, Quoc~V Le, Denny Zhou, et~al.
\newblock Chain-of-thought prompting elicits reasoning in large language models.
\newblock \emph{NPIS}, 35:\penalty0 24824--24837, 2022{\natexlab{b}}.

\bibitem[Zeng et~al.(2022)Zeng, Liu, and Zhou]{zeng2022tag}
Jiandian Zeng, Tianyi Liu, and Jiantao Zhou.
\newblock Tag-assisted multimodal sentiment analysis under uncertain missing modalities.
\newblock In \emph{SIGIR}, pages 1545--1554, 2022.

\bibitem[Zhang et~al.(2022)Zhang, Chu, Ma, Zhu, Wang, Wang, and Zhao]{zhang2022m3care}
Chaohe Zhang, Xu Chu, Liantao Ma, Yinghao Zhu, Yasha Wang, Jiangtao Wang, and Junfeng Zhao.
\newblock M3care: Learning with missing modalities in multimodal healthcare data.
\newblock In \emph{SIGKDD}, pages 2418--2428, 2022.

\bibitem[Zhang et~al.(2023)Zhang, Rao, and Agrawala]{zhang2023adding}
Lvmin Zhang, Anyi Rao, and Maneesh Agrawala.
\newblock Adding conditional control to text-to-image diffusion models.
\newblock In \emph{CVPR}, pages 3836--3847, 2023.

\bibitem[Zhao et~al.(2021)Zhao, Li, and Jin]{zhao2021missing}
Jinming Zhao, Ruichen Li, and Qin Jin.
\newblock Missing modality imagination network for emotion recognition with uncertain missing modalities.
\newblock In \emph{ACL}, pages 2608--2618, 2021.

\end{thebibliography}
}


\end{document}